\documentclass[letterpaper]{article}
\usepackage{aaai25} 
\usepackage{times} 
\usepackage{helvet} 
\usepackage{courier} 
\usepackage[hyphens]{url}
\usepackage{graphicx} 
\usepackage{natbib} 
\usepackage{caption}
\usepackage{mathptmx}
\usepackage{algorithm}
\usepackage{algorithmic}

\usepackage{newfloat}
\usepackage{listings}

\usepackage{booktabs}
\usepackage[table]{xcolor}
\usepackage{array}
\usepackage{multirow}

\usepackage{pdfpages}

\usepackage{amsmath}
\usepackage{tcolorbox}
\usepackage{cuted}

\newcommand\blfootnote[1]{%
  \begingroup
  \renewcommand\thefootnote{}\footnote{#1}%
  \addtocounter{footnote}{-1}%
  \endgroup
}

\DeclareCaptionStyle{ruled}{labelfont=normalfont,labelsep=colon,strut=off} 
\floatstyle{ruled}
\newfloat{listing}{tb}{lst}{}
\floatname{listing}{Listing}

\title{MuMA-ToM: Multi-modal Multi-Agent Theory of Mind}
\author {
    Haojun Shi\textsuperscript{\rm 1*},
    Suyu Ye\textsuperscript{\rm 1*},
    Xinyu Fang\textsuperscript{\rm 1},
    Chuanyang Jin\textsuperscript{\rm 1},
    Leyla Isik\textsuperscript{\rm 1},
    Yen-Ling Kuo\textsuperscript{\rm 2},\\
    Tianmin Shu\textsuperscript{\rm 1}
}
\affiliations {
    \textsuperscript{\rm 1}Johns Hopkins University, \\
    \textsuperscript{\rm 2}University of Virginia\\
    \{hshi33, sye10, xfang21, cjin33, lisik, tianmin.shu\}@jhu.edu, ylkuo@virginia.edu
}

\begin{document}
\maketitle

\blfootnote{* Equal contribution.}

\begin{abstract}
Understanding people's social interactions in complex real-world scenarios often relies on intricate mental reasoning. To truly understand how and why people interact with one another, we must infer the underlying mental states that give rise to the social interactions, i.e., Theory of Mind reasoning in multi-agent interactions. Additionally, social interactions are often multi-modal -- we can watch people's actions, hear their conversations, and/or read about their past behaviors. For AI systems to successfully and safely interact with people in real-world environments, they also need to understand people's mental states as well as their inferences about each other's mental states based on multi-modal information about their interactions. For this, we introduce MuMA-ToM, a Multi-modal Multi-Agent Theory of Mind benchmark. MuMA-ToM is the first multi-modal Theory of Mind benchmark that evaluates mental reasoning in embodied multi-agent interactions. In MuMA-ToM, we provide video and text descriptions of people's multi-modal behavior in realistic household environments. Based on the context, we then ask questions about people's goals, beliefs, and beliefs about others' goals. We validated MuMA-ToM in a human experiment and provided a human baseline. We also proposed a novel multi-modal, multi-agent ToM model, LIMP (Language model-based Inverse Multi-agent Planning). Our experimental results show that LIMP significantly outperforms state-of-the-art methods, including large multi-modal models (e.g., GPT-4o, Gemini-1.5 Pro) and a recent multi-modal ToM model, BIP-ALM.
\end{abstract}

\begin{links}
\link{Code and data}{ https://scai.cs.jhu.edu/projects/MuMA-ToM/}
\end{links}

\begin{figure*}[!htbp]
    \centering
    \includegraphics[trim = 0cm 11cm 0cm 0cm, clip, width=\textwidth]{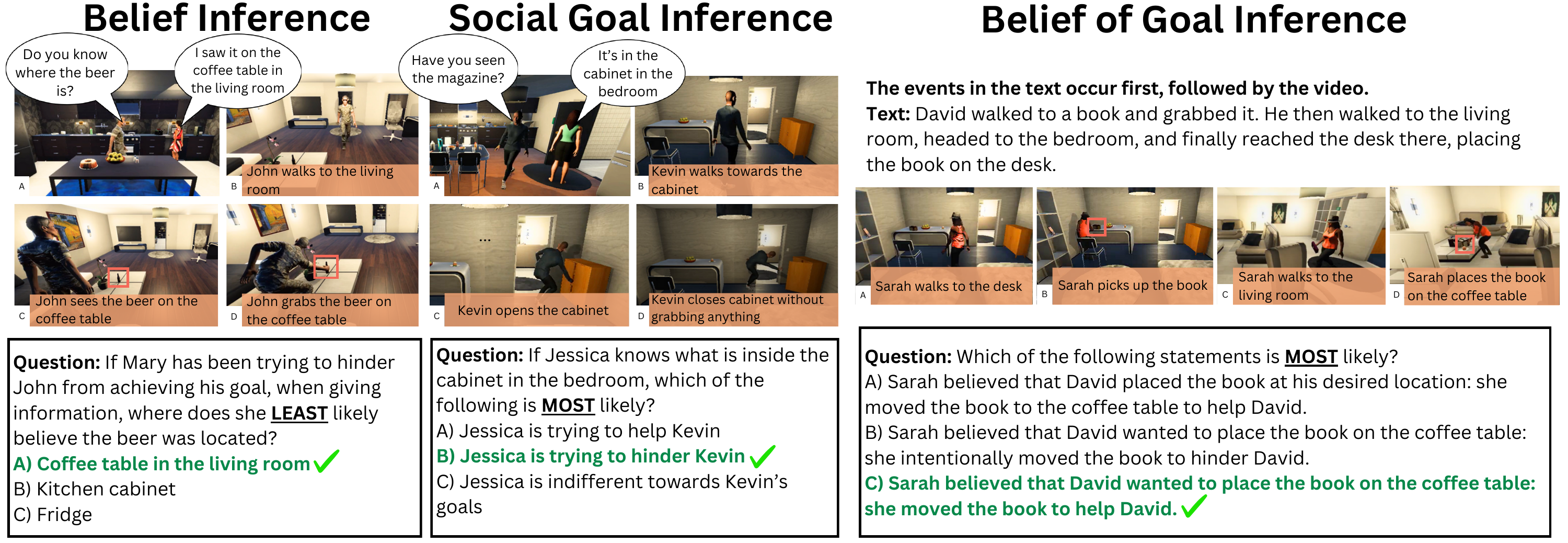} 
    
    \caption{Example questions for each question type. We provide keyframes for the video in each example. The conversations in the chat bubbles are provided as subtitles and shown as part of the multi-modal inputs when viewing the video. Note that the captions on the bottom of the frames are for illustrative purposes only and are not shown in the videos. The checkmarks indicate the correct answers. We provide the videos and text for the examples in the supplementary material.}
    \label{fig:sample_questions}
\end{figure*}

\section{Introduction}
Humans live in a social world; we not only engage in social interactions ourselves but can also understand other people's social interactions. Studies in Developmental Psychology have indicated that the ability to understand different kinds of social interactions develops early and is one of the bases for more sophisticated social skills developed later in life \cite{denham2003preschool, wellman2001meta, hamlin2007social}. Crucially, understanding social interactions goes beyond action recognition. We often need to reason about \textit{why} people interact with one another in a certain manner. We can achieve this by inferring people's mental states as well as how they reason about one another's mental states, i.e., multi-agent Theory of Mind (ToM) reasoning. For instance, if Alice puts away a book on Bob's desk, she may be trying to clean up or hide the book, depending on both her social goal (helping or hindering) and where she believes Bob wants the book (belief of other's goal). As an observer, it may be difficult to disambiguate between these scenarios. However, if we had heard Bob asking Alice where the book was before, we would confidently infer that Alice wanted to hinder Bob. Such multi-modal, multi-agent Theory of Mind abilities are not only crucial for humans but also for AI systems that are deployed in human living environments, such as assistive robots. Without a robust understanding of people's mental states in complex social interactions, AI systems may cause detrimental errors in their interactions with people.

Despite the recent advances in evaluating and engineering machine Theory of Mind, prior works have not adequately addressed the challenge of Theory of Mind reasoning in multi-modal social interactions. First, common Theory of Mind benchmarks \cite{Gordon2016CommonsenseIO, gandhi2021baby, shu2021agent, kosinski2023theory, jin2024mmtom} have only focused on individuals' mental states. However, there are other important aspects of multi-agent mental reasoning, including social goals (e.g., helping, hindering) and beliefs about others' goals. Second, there has not been a multi-modal social interaction dataset designed for systematic Theory of Mind reasoning evaluation. The only prior multi-modal Theory of Mind dataset is MMToM-QA, which solely focuses on single-agent activities. Text-only benchmarks such as Hi-TOM \cite{wu-etal-2023-hi} feature multi-agent events, but lack visual inputs. Thus, it remains unclear how we can evaluate the multi-modal multi-agent Theory of Mind capacity in machine learning models.

To address these challenges, we introduce a new Theory of Mind benchmark, MuMA-ToM (Multi-modal Multi-Agent Theory of Mind benchmark). MuMA-ToM includes a large set of question-answering trials.  As summarized in Figure~\ref{fig:sample_questions}, questions in MuMA-ToM are organized into three categories: (1) belief inference, (2) social goal inference, and (3) belief of goal inference. In each trial, there is a multi-agent event in a household environment depicted by video and text. As shown in Figure~\ref{fig:sample_questions}, in some trials, text may show a conversation between two agents; in other trials, text may describe a part of an event that is not depicted in the video. Based on the multi-modal inputs, there will be a question about agents' mental states. We evaluated both humans and state-of-the-art multi-modal models on MuMA-ToM. While humans can achieve near-perfect performance, baselines all fail to robustly infer the mental states based on the multi-modal context.

To bridge the gap between human ToM and machine ToM, we propose a novel multi-modal multi-agent Theory of Mind method -- LIMP (Language model-based Inverse Multi-modal Planning). Inspired by a recent method, BIP-ALM, proposed by \cite{jin2024mmtom}, LIMP incorporates language models as components for inverse planning. Unlike BIP-ALM, LIMP (1) introduces multi-agent planning with two-level reasoning, (2) eliminates the need for manually defined symbolic representations for a better generality, and (3) can leverage any pretrained LLMs whereas BIP-ALM requires LLMs finetuned on symbolic representations. Experimental results demonstrate that LIMP significantly outperforms baselines.

In sum, our contribution includes (1) the first benchmark on multi-modal multi-agent Theory of Mind reasoning, (2) a human experiment validating the benchmark and providing a human baseline, (3) a systematic evaluation of state-of-the-art large multi-modal models (LMMs), and (4) a novel multi-modal multi-agent ToM method combing inverse multi-agent planning and language models.

\section{Related Works}

Single-agent ToM benchmarks \cite{Gordon2016CommonsenseIO, gandhi2021baby, shu2021agent, kosinski2023theory, jin2024mmtom} have extensively tested concepts like belief, goal, preferences, constraints, and rationality. Multi-agent benchmarks are typically built based on the classic Sally-Anne test \cite{baron1985does} for false beliefs and higher-order beliefs \cite{le2019revisiting, he2023hi, xu2024opentom, soubki2024views}. There have also been multi-agent benchmarks that focus on a single agent's beliefs \& intentions in complex conversations or interactions \cite{kim2023fantom, chen2024tombenchbenchmarkingtheorymind, chan2024negotiationtom, sabour2024emobenchevaluatingemotionalintelligence}. In these benchmarks, other agents are usually present to add context or complexity, but there are no questions about inter-agent relationships. Prior works on testing social relationship understanding \cite{netanyahu2021phase, li2024infant} rely on simple animations, which lack the realism of embodied human interactions. Most existing ToM benchmarks have only either text or video. The only exception is MMToM-QA \cite{jin2024mmtom}, which has single-agent activities depicted in video and text. Our MuMA-ToM benchmark features two agents interacting in an embodied household environment, with both text and video as multi-modal inputs, and includes questions that test the agents' social intentions and their reasoning about each other's mental states.

\subsubsection{Multi-Modal Benchmarks.} Given the recent advances in LLMs, there has been increasing interest in developing multi-modal QA benchmarks. Most of these benchmarks focus on models' ability to fuse information from multiple modalities, where answers are directly retrievable without complex reasoning  \cite{li2023m3itlargescaledatasetmultimodal, sanders2023multivent, li2023seed, ying2024mmt, tang2024mtvqa, pandya2024ntsebenchcognitivereasoningbenchmark}. A recent benchmark, Perception Test \cite{patraucean2024perception}, evaluates physical reasoning such as predicting world states and explaining counterfactual facts. But it differs from ToM reasoning. Pipelines for generating multi-modal datasets, SEED-story \cite{yang2024seedstory} and TaskMeAnything \cite{zhang2024task}, also do not evaluate ToM reasoning. MMToM-QA \cite{jin2024mmtom}, a recent multi-modal ToM benchmark, evaluates ToM with multi-modal inputs about single-agent behaviors. Unlike MMToM-QA, our benchmark includes multi-agent interactions and evaluates models' understanding of mental state reasoning in multi-modal social interactions.

\subsubsection{Machine Theory of Mind.} Traditional approaches to Theory of Mind reasoning fall into two categories: end-to-end training \cite{rabinowitz2018machinetheorymind, Han_Gmytrasiewicz_2019} and Bayesian Inverse Planning \cite{baker2017rational, zhi2020online, stacy2024bayesian}. There have been works on neural amortized inference that combine these two methods for efficient and robust ToM inference in visual domains \cite{jha2024neural, puig2023nopa}. Recently, LLMs demonstrated some ToM reasoning capabilities \cite{kosinski2023theory, bubeck2023sparks}, but their ToM reasoning is still brittle \cite{verma2024theory, amirizaniani2024llms, ullman2023largelanguagemodelsfail, sclar2023minding, ivanova2024elementsworldknowledgeewok}. Approaches using prompt engineering have been proposed to enhance the ToM capacities in LLMs for text-based QAs \cite{wilf2023thinktwiceperspectivetakingimproves, sclar-etal-2023-minding}. \citet{jin2024mmtom} proposed, BIP-ALM, for multi-modal ToM. BIP-ALM first extracts and fuses symbolic representations from multi-modal inputs and then combines a language model and Bayesian inverse planning to conduct ToM reasoning based on the symbolic representations. While achieving promising results on MMToM-QA, BIP-ALM lacks multi-agent reasoning capacity and requires finetuning a language model on hand-designed symbols. Our LIMP model builds on BIP-ALM and introduces key improvements including multi-agent planning and general, domain-invariant representations. 


\section{MuMA-ToM Benchmark}

\subsection{General Structure}
The benchmark consists of 225 multi-modal social interactions between two agents. There are 900 multi-choice questions based on these social interactions. Each question depicts a social interaction in video and text jointly. As shown in Figure~\ref{fig:sample_questions}, the text may show a conversation between the agents or a part of the event, and the video shows the complementary part of the event. Given the multi-modal inputs, the questions are designed to assess the understanding of agents' mental states during these interactions, probing three main concepts: (1) beliefs, (2) social goals, and (3) beliefs of others' goals. Each concept has 300 questions. We also created a training set consisting of 1,030 videos annotated with the agents' actions and goals. The training set does not provide example questions. It is intended for a model to learn about typical multi-agent household activities.

\subsection {Question Types}

As identified in prior works in cognitive science \citep{ullman2009help,shu2020adventures} and multi-agent planning \cite{Gmytrasiewicz_2005,tejwani2021social}, there are three mental variables that are crucial to ToM reasoning in multi-agent interactions: an agent's belief of the physical state, its social goal, and its belief of other agents' goals. Therefore, we design three types of questions in our benchmark corresponding to the three mental variables: belief inference, social goal inference, and belief of goal inference. Each type of question asks about the corresponding mental variable of one of the agents. Among the three options, we make sure that there is always one option that is clearly the most likely to be correct.

One of the challenges in designing these three types of questions is that given an interaction, multiple combinations of these mental variables could be equally possible. For instance, if we see that Alice's actions prevent Bob from reaching his goal, it could be because Alice is hindering Bob, knowing Bob's true intent; or she may try to help Bob but has a false belief of Bob's goal and ends up accidentally hindering Bob. To address the challenge of large hypothesis space, we always ask a question about a mental variable conditioned on explicitly provided assumptions about the other two mental variables. For instance, as shown in the example question of the belief inference in Figure~\ref{fig:sample_questions}, the goal of John can be inferred from his question about where the beer is (the goal is getting beer), and Mary should be aware of this as she answered John's question; the social goal of Mary is unclear, therefore the question provides a hypothetical social goal, hindering, as the condition. The remaining mental variable of Mary, her belief of the physical state can be clearly inferred given her social goal and her belief of John's goal.

We explain the design of each question type as follows.

\textbf{Belief Inference.}
These questions focus on inferring a person's belief about the physical state based on their utterance and social goal. The person may have a \textit{true belief} or \textit{false belief} about the location of the object, which can be inferred when we constrain their social goal to be helping or hindering. In the example depicted in Figure~\ref{fig:sample_questions}, John asks Mary where he can find the beer. Mary suggests the coffee table, which turns out to be the correct location, as John successfully finds the beer there. This could be interpreted in two ways: (1) Mary helps John, genuinely believing the beer is on the coffee table, or (2) Mary accidentally helps John while intending to mislead him, mistakenly believing that the beer isn't on the coffee table. To answer correctly, a model needs to understand: (1) Mary knows John's goal (from their conversation), (2) John follows Mary's directions (from their conversation and his actions afterward in the video), and (3) John achieves his goal by following Mary's directions (as shown in the video). We balance true and false beliefs in the ground-truth answers. 

\textbf{Social Goal Inference.}
In these questions, we ask about a person's social goal. Specifically, we consider helping, hindering, or acting independently as the three possible social goal categories, which are also the common social goal types in physically grounded social interaction reasoning studied by prior works in cognitive science \cite{hamlin2007social,ullman2009help,shu2020adventures,malik2023relational}. The example in Figure~\ref{fig:sample_questions} shows an interaction similar to the one in the example for belief inference questions. In this particular example, Jessica misleads Kevin to the cabinet where there is no magazine inside. In the question, we assume that Jessica does indeed know the true state, and therefore, one should infer that Jessica is trying to hinder Kevin. To achieve this correct inference, a model needs to focus on (1) how Jessica infers Kevin's goal (from the conversation), (2) how Kevin searches the room after the conversation (from both the conversation and the video following the conversation), and (3) whether Kevin can find his goal object at the location suggested by Jessica (from the video). We balance cooperative and adversarial behaviors for the ground-truth answers. 

\textbf{Belief of Goal Inference.}
Belief of goal inference asks a model of how one person thinks about another person's goal given the context. In each option for a question of this type, we always pair the belief of another person's goal with the corresponding social goal to minimize ambiguity. For instance, in the interaction for the example question of belief of goal inference in Figure~\ref{fig:sample_questions}, Sarah moves the book to the coffee table after David places it on the desk. However, it is unclear whether Sarah is aware that David places the book there and whether Sarah thinks that David wants to keep the book on the desk. If Sarah were trying to help David, as assumed in the correct option, she would have believed that David wanted the book on the coffee table instead. In this case, as a third-person observer, we may not be certain of David's true intent, but we can still infer Sarah's belief of David's goal given that her social goal is helping him. For this type, half of the questions have a true belief of goal as the correct answer, and the other half have a false belief of goal as the correct answer.

\subsection{Multi-modal Information}
In MMToM-QA, the only prior multi-modal ToM QA benchmark, each modality covered \textit{all} aspects of the human activity and environment, making it difficult to discern what information could be extracted from one modality but not the other. Our benchmark aims to provide clearly separate information accessible only through one modality, allowing us to understand precisely how a model needs to fuse multi-modal inputs to answer each question.

As illustrated in Figure~\ref{fig:sample_questions}, there are two main ways in which multi-modal information must be integrated. First, if there are conversations between two agents, the model must understand the exchanged information and how it impacts each person's mental state, including any changes in their beliefs about each other. The model must also observe actions and outcomes, connecting them to the conversation to reason further about mental states. Note that conversations can occur at any point in the video. Second, for interactions without verbal communication, we provide part of the event in text and the remaining part in video. Specifically, we either describe the first half in text and show the second part in video or show the first part in video and describe the second half in text. These two designs are randomly sampled to describe interactions jointly in video and text.

\subsection{Procedural Generation}
We use a multi-agent household simulator, VirtualHome \cite{puig2018virtualhome, puig2020watchandhelp}, to procedurally synthesize social interactions between two agents. For each interaction, we sample an environment and goals for the agents. We consider three general social scenarios: an agent is trying to help another agent, an agent is trying to hinder another agent, and two agents are acting independently. Agents only have partial observations and do not know each others' goals. They can optionally talk to each other. We leverage a recent method proposed by \citet{ying2024gomaproactiveembodiedcooperative}---Goal-Oriented Mental Alignment (GOMA)----to generate action plans as well as verbal communication. GOMA combines hierarchical planning, goal inference, and large language models (LLMs) to generate multi-modal interactions between embodied agents.  Prior work \cite{puig2020watchandhelp} has demonstrated that activities synthesized in VirtualHome indeed resemble real-world human activities. We provide more details on the procedural generation in the supplementary material.

\begin{figure*}[t!]
    \centering
 
        \includegraphics[trim=0cm 9.7cm 0cm 0cm, clip, width=\textwidth]{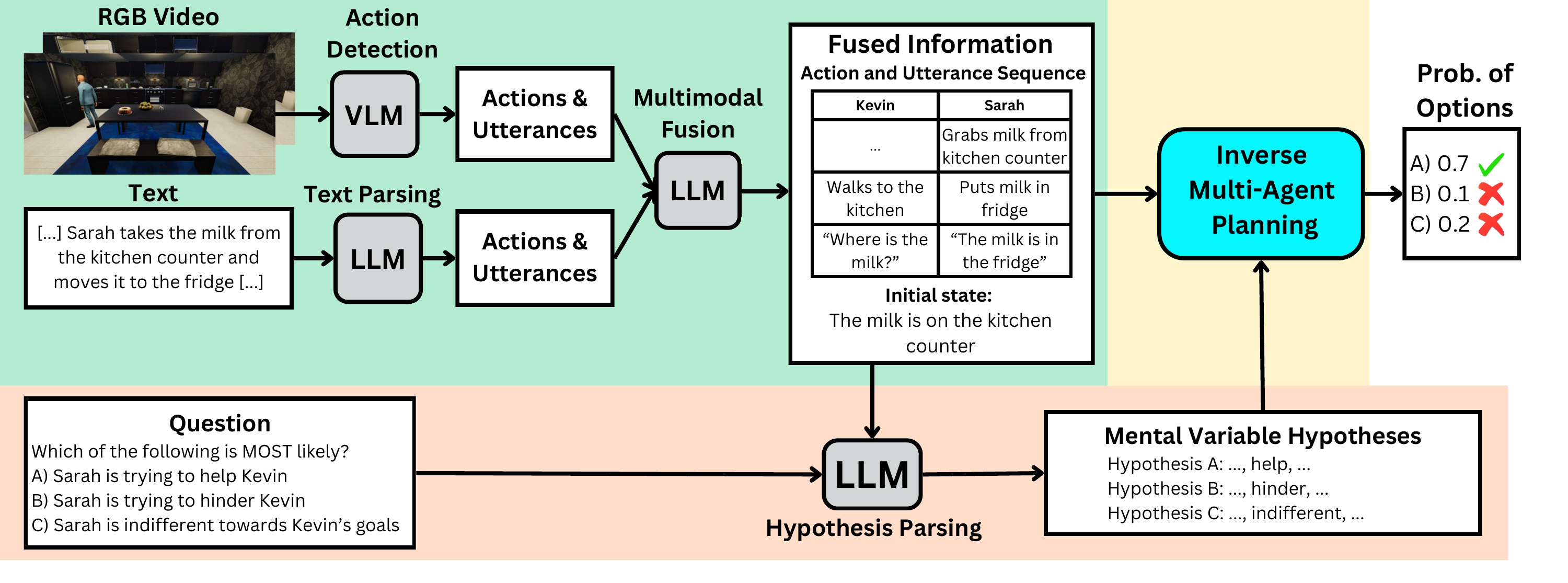}

        \caption{Overview of LIMP. LIMP has three components: (1) the multi-modal information fusion module extracts and fuses information from vision and text; (2) the hypothesis parsing module generates hypothetical values for the three mental variables given the question and the fused information; and (3) the inverse multi-agent planning module assesses the probabilities of each option given the hypothetical mental variables and the multi-modal agent behavior described in the fused information.} 
\label{fig:LIMP}
\end{figure*}

\section{Our Model}

\subsection{Formulation}

To model social interactions between two agents, $i$ and $j$, and the recursive mental reasoning between them, we adopt an Interactive Partially Observable Markov Decision Processes (I-POMDP) formulation \cite{Gmytrasiewicz_2005}. We define \( s^t \) as the state,  \( a_i^t \) and \( a_j^t \) as agents' actions, and  \( u_i^t \) and \( u_j^t \) as agents' utterances at time $t$. Each agent maintains its own beliefs \( b_i^t \) and \( b_j^t \), as well as goals \( g_i \) and \( g_j \). To capture recursive reasoning, we define interactive states for the agents, denoted as \( is_{i, \ell} \) and \( is_{j, \ell} \) at level \(\ell\). From the perspective of agent \(i\), its interactive state at each level is defined as follows (we consider the first two levels in this work):
\begin{itemize}
    \item \textbf{Level 0:} \( is_{i, 0} = s \)
    \item \textbf{Level 1:} \( is_{i, 1} = (s, b_{j, 0}, g_j) \)  (where \( b_{j, 0} \) is a distribution over agent \( j \)'s level 0 interactive state, \( is_{j, 0} \))
    \item ...
\end{itemize}
Given the belief of interactive state $b(is_{i,1})$, an agent's action policy will be $\pi(a_i | is_{i,1}, g_i)$, and its utterance policy will be $\pi(u_i | is_{i,1}, g_i)$.

\subsection{Overview}

Previous works on Inverse Multi-agent Planning (IMP) \cite{ullman2009help,netanyahu2021phase} have demonstrated that IMP can robustly infer agents' mental states in social interactions. However, these methods rely on manually crafted planners and are limited to simple visual scenarios, such as 2D grid worlds. \citet{jin2024mmtom} introduced the BIP-ALM model, which leverages language models for inverse planning to achieve single-agent Theory of Mind reasoning in complex, realistic settings. Inspired by BIP-ALM, we propose a novel method, Language model-based Inverse Multi-agent Planning (LIMP), to combine IMP and language models for robust multi-agent Theory of Mind reasoning based on multi-modal inputs.

\begin{figure*}[h!]
    \centering

    \includegraphics[trim=2cm 5.5cm 1cm 5cm, clip, width=0.8\textwidth]{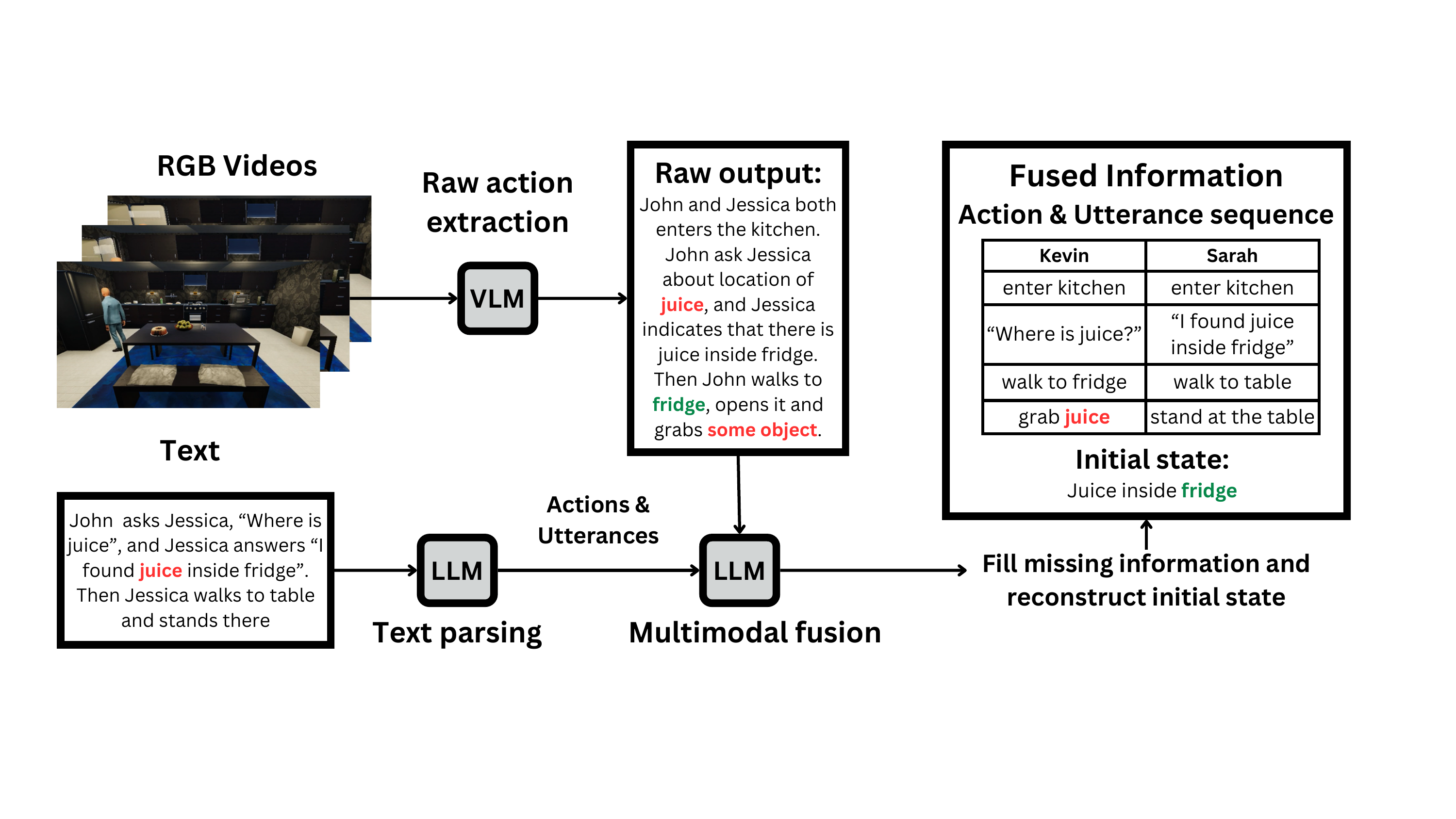}
    \caption{Illustration of the multi-modal information fusion in LIMP. It fills in missing information based on the context and recovers the initial state from agents' actions.}
   \label{fig:error_correction}
\end{figure*}

As illustrated in Figure~\ref{fig:LIMP}, LIMP consists of three key components: multi-modal information fusion, hypothesis parsing, and inverse multi-agent planning. Compared to BIP-ALM, our approach offers several improvements. First, while BIP-ALM is limited to single-agent scenarios, LIMP identifies three mental variables crucial to understanding multi-agent interactions—belief, social goal, and belief of goal. We then implement multi-agent planning based on these variables to reason about multi-modal social interactions. Second, BIP-ALM relies on hand-designed symbolic representations and requires finetuning language models on these representations. In contrast, LIMP uses natural language to represent states, actions, and utterances, eliminating the need for finetuning and enhancing generalizability across domains. Finally, LIMP's multi-modal information fusion module can fill in missing information from visual perception using contextual cues from text or action sequences (Figure~\ref{fig:error_correction}), a capability absent in BIP-ALM. We discuss the details of each component in the remaining section.

\begin{figure*}[t!]
    \centering
        \includegraphics[trim=1cm 1cm 1cm 0cm, clip, width=0.8\textwidth]{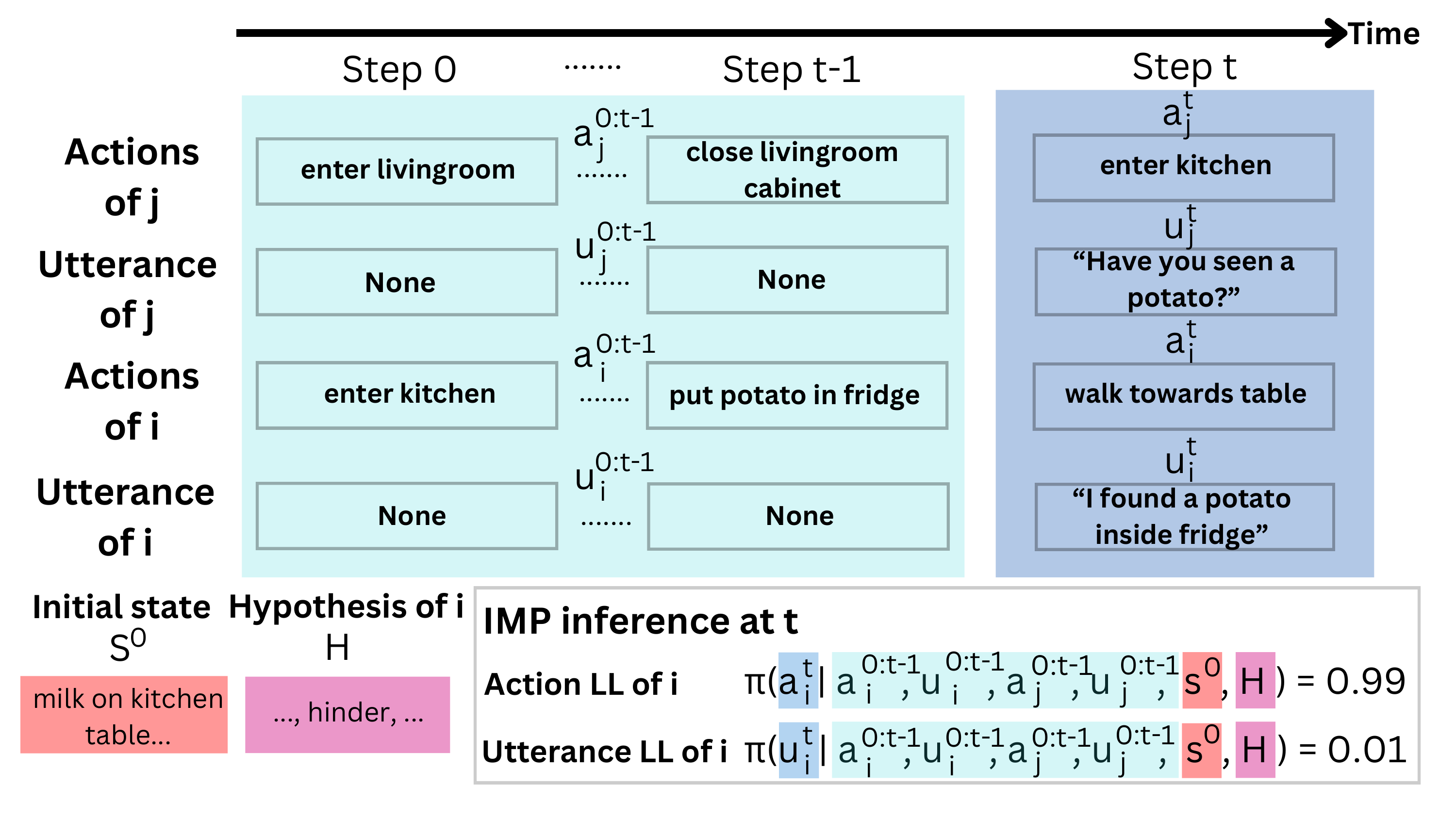}
        \caption{Illustration for inverse multi-agent planning. We estimate the action and utterance likelihood of agent $i$ at each step $t$ given the past actions and utterances of both agents from step 0 to step $t-1$, the initial state $s^0$, and the hypothesis $H$. LL in the figure stands for likelihood.}
 \label{fig:IMP}
\end{figure*}

\subsection{Multi-modal Information Fusion}

We use a vision-language model (VLM) to extract the actions and utterances of each person depicted in the video. Given text, we use an LLM to extract the actions and utterances of each person. We then fuse the extracted information to form the initial state and the complete sequences of actions and utterances using an LLM as follows.

Unlike MMToM-QA, our benchmark does not provide a text description of the full state, as such descriptions are rarely provided in real-world applications. However, as objects may be occluded or too small to detect even for humans, inferring the state directly from the RGB videos could be difficult. Instead, we prompt an LLM with the inferred actions and utterances of both agents to infer the part of the initial state relevant to the activity. For example, if Alice grabs a carrot from the fridge, and moves it to the kitchen table, we can infer that the carrot was originally in the fridge. Using this method, the reconstructed initial state will only consider objects relevant to human actions and utterances. This simplifies the context and can consequently improve the accuracy of the inference. Given the initial state and the action sequences, we can infer the state at each step.

There is often missing information in the visual perception results. For instance, as shown in Figure~\ref{fig:error_correction}, the VLM did not recognize the object the person grabbed and produced an ambiguous action -- ``grabs some object.'' This is also common to people, as the object picked up by the person is often occluded. However, we can still infer that the object is likely juice based on the context provided in the text. To emulate such ability, we leverage an LLM to fuse information extracted from video and text, which infers the information missing from visual perception based on the complementary information described in the text. In this work, we use Gemini 1.5 Pro for the VLM and GPT-4o for the LLM as they produce the best results.

\subsection{Hypothesis Parsing}

To answer the question about a person's mental state in a social interaction, LIMP will parse relevant hypotheses of all mental variables of that person (agent $i$) -- belief of state $b(s)$, social goal $g_i$, and belief of other agent's goal $b(g_j)$. For this, we prompt GPT-4o with the initial state and question text to generate a reasonable hypothesis of the three mental variables for each option, $H = \langle b(s), g_i, b(g_j)\rangle$.

\subsection{Inverse Multi-Agent Planning}

Given the fused information from multi-modal inputs and the parsed hypotheses, inverse multi-agent planning conducts Bayesian inference over a person's mental state by evaluating the likelihood of actions and utterances given each hypothesis. Following the I-POMDP formulation, we define this probabilistic inference as follows:
\begin{align}
 &P(H \mid a_{i}^{0:T}, u_{i}^{0:T}, a_{j}^{0:T}, u_{j}^{0:T}, s^{0}) \nonumber\\
\propto& P(H) \prod_{t=1}^{T} \pi(a_{i}^t\mid a_i^{0:t-1},u_i^{0:t-1},a_j^{0:t-1},u_j^{0:t-1}, s^0, H)\nonumber\\  
&\cdot \prod_{t=1}^{T} \pi(u_{i}^t\mid a_i^{0:t-1},u_i^{0:t-1},a_j^{0:t-1},u_j^{0:t-1}, s^0, H),
\end{align}
where the action policy and the utterance policy can be estimated by the log probabilities of the prompt completion by a language model for each time step $t$. Note that in the standard policy definitions in I-POMDP, we need agent $i$'s belief of agent $j$'s belief of the state at each step. This, however, is difficult to explicitly estimate. Instead, in this work, we consider past actions and utterances of all agents as part of the condition of the policies to avoid the explicit belief of belief inference. We prompt an LLM with the hypothesis, the initial state, and the previous actions and utterances of both agents to estimate the action and utterance policies. In this way, we do not need to implement domain-specific planning, which can be extremely challenging and slow for multi-agent interactions with both physical actions and verbal communication. In this work, we use GPT-4o for the LLM. We find that GPT-4o can accurately estimate the action and utterance policies based on the given condition. 

Figure~\ref{fig:IMP} illustrates how IMP evaluates the action and utterance likelihood at one time step. Given the condition, the LLM estimates that it is likely that agent $i$ will take the observed action (``walk towards table'') but is unlikely to say ``I found a potato inside fridge'' as it is inconsistent with the social goal of hindering agent $j$ (agent $i$ had just put a potato in the fridge before the conversation).

\section{Experiments}

\subsection{Human Experiment} We recruited 18 participants (mean age = 36.0; 10 female) from Prolific to answer 90 questions randomly sampled from the benchmark. Each question received responses from 3 participants. The experiment was approved by an institutional review board.

\subsection{Baselines}
We evaluated our benchmark on state-of-the-art LMMs. For models capable of processing video input, the entire video was provided. For models without video input capabilities, we uniformly sample one frame every 20 frames from the video episode as input. We evaluated \textbf{GPT-4o} \cite{OpenAI2023GPT4TR}, \textbf{Llava 1.6} \cite{liu2023llava}, \textbf{Gemini 1.5} \cite{reid2024gemini}, \textbf{InternVL2} \cite{chen2023internvl} and \textbf{VideoLlama 2} \cite{damonlpsg2024videollama2}. We evaluated the latest version of each LMM at the time of submission. For \textbf{LIMP}, we use Gemini 1.5 Pro as the VLM and GPT-4o as the LLM. Finally, we evaluated \textbf{BIP-ALM} using finetuned Llama 2\cite{jin2024mmtom}, the best-performing model on a prior multi-modal ToM benchmark, MMToM-QA. For this evaluation, we used the original \textbf{BIP-ALM} model, which was fine-tuned on the MMToM-QA dataset. More details of the experiments are provided in the supplementary material.

\begin{figure}[t!]
    \centering
    \includegraphics[trim=3.6cm 16.9cm 6.3cm 3.1cm, clip, width=0.38\textwidth]{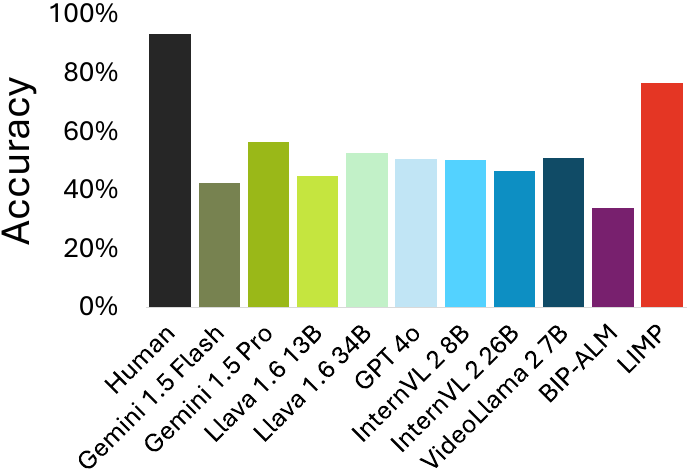}
    \caption{Human and model performance on MuMA-ToM.}
    \label{fig:performance_graph}
\end{figure}

\subsection{Results}
We report the human and model performance in Figure~\ref{fig:performance_graph} and Table~\ref{tab:results}. Human participants achieved almost perfect accuracy across all questions, with 98.9\% of the correct answers having majority agreement. The overall performance averaged across individual participants is $93.5\%$. The slightly lower performance on social goal inference (94.4\%) and belief of goal inference (87.1\%) indicates these questions are more challenging and require greater focus.

\begin{table}[t!]
  \begin{center}
    \begin{small}
    \begin{tabular}{l|p{1cm}|p{1cm}|p{1cm}|p{1cm}}
    \toprule
      \textbf{Method} & \textbf{Belief} & \textbf{Social Goal} & \textbf{Belief of Goal} & \textbf{All}\\
    \hline
    \rowcolor{gray!20}
    Human & 98.9 & 94.4 & 87.1 & 93.5 \\
    Gemini 1.5 Flash & 53.9 & 33.0 & 41.4 & 42.7 \\
    Gemini 1.5 Pro & 78.9 & 43.9 & 46.9 & 56.4 \\
    Llava 1.6 13B & 70.2 & 43.2 & 17.9 & 43.7 \\
    Llava 1.6 34B & \textbf{93.6} & 37.2 & 27.5 & 52.8 \\
    GPT-4o & 67.9 & 39.6 & 44.4 & 50.6 \\
    InternVL 2 8B & 62.2 & 44.6 & 45.1 & 50.6 \\
    InternVL 2 26B & 59.3 & 44.9 & 35.5 & 46.6 \\
    VideoLlama 2 7B & 70.1 & 45.6 & 37.7 & 51.1 \\
    BIP-ALM & 41.2 & 34.1 & 30.6 & 33.9 \\
    LIMP & 93.4 & \textbf{67.7} & \textbf{68.7} & \textbf{76.6} \\
    \bottomrule
    \end{tabular}
    \end{small}
    \caption{Human and model performance for different question types as well as for all questions.}
    \label{tab:results}
  \end{center}

\end{table}

All LMM baselines performed poorly on MuMA-ToM, indicating a substantial gap between machine and human ToM. Among the three question types, belief inference is the easiest for LMMs. In particular, Llava 34B achieved the highest accuracy for belief inference. However, all LMMs struggle with the more challenging social goal inference and belief of goal inference questions. The best-performing LMM model overall is Gemini 1.5 Pro, with an accuracy of 56.4\%. Notably, BIP-ALM had an accuracy of 33.9\%, indicating its inability to understand multi-agent interactions. This is because BIP-ALM only relies on single-agent goals and beliefs for inverse planning but does not consider social goals and beliefs of others' goals. Additionally, BIP-ALM assumes certain symbolic representations, which also limits its generality. Our LIMP model significantly outperforms allstate-of-the-art models on our benchmark, with an overall accuracy of 76.6\%. Critically, by using GPT-4o to estimate action and utterance likelihood for inverse multi-agent planning, LIMP can achieve much better performance than GPT-4o itself. There is still a gap between the best model performance and human performance, highlighting the need for further studies.

We finetuned VideoLlama 2 on a video instruction dataset created from our training set. The finetuned model did not perform better, suggesting that common finetuning approaches for LMMs do not improve their ToM capacities. We also evaluated the performance of LMMs with chain-of-shot prompting \cite{Kojima2022LargeLM} and found no improvement (Table 2 in the supplementary material). 

\section{Discussion}

\textbf{Why do LMMs perform poorly?} There are two sources of systematic errors for LMMs. First, LMMs struggle with understanding multi-modal behavior in complex social situations, often failing to distinguish between deliberate hindering and failed attempts to help due to incorrect beliefs. Most models can solve belief inference tasks where helping is the assumed social goal. However, they consistently struggle in scenarios where hindering is the assumed social goal. (e.g., ``If Mary is trying to hinder Jack, where does she least likely believe...?") The failure to understand adversarial behaviors is even more prominent in social goal inference and belief of goal inference. Second, LMMs often fail to correctly interpret visual inputs, such as when an object is too small or is occluded when the agent is picking it up, leading to incorrect conclusions about the agent's actions. While humans are able to use contextual clues to infer what the object might be, VLMs struggle with this task. These errors in recognizing crucial actions likely contribute significantly to their overall poor performance on our benchmark.

\textbf{Why does LIMP outperform the best LMMs?} 
LIMP overcomes the two aforementioned weaknesses of LMMs -- the inability to recognize multi-modal behavior under various social goals and the sensitivity to noisy visual perception. First, while LLMs struggle with direct ToM reasoning, they excel at the forward generation of multi-modal behavior given mental states. For example, it is much harder for an LLM to correctly infer whether an agent is hindering another by lying than it is for the model to generate a lie based on the agent's belief and social goal. Such multi-modal behavior generation ability enables LIMP to estimate the action and utterance likelihood, identifying the key actions and/or utterances that reveal the true mental state of an agent (as shown in Figure~\ref{fig:IMP}). Second, when a VLM fails to recognize the exact object that the agent is interacting with, LIMP can fill in this missing information with context from the text input. We evaluated the action accuracy by using semantic similarity and found that this approach increases inferred action accuracy from 54.4\% to 86.6\%. As a result, LIMP is able to perform inference on much more accurate information.

Further ablation studies highlight the critical role of individual LIMP components in its performance. GPT-4o provided with ground truth actions achieves an accuracy of only 53.2\%, and LIMP without inverse planning achieves an accuracy of 55.3\%. Both are much lower than LIMP's performance of 76.6\%, showing the importance of inverse planning.

\textbf{How general is LIMP?}
Prior inverse planning models, including BIP-ALM, all require handcrafted representations for specific domains. LIMP, however, represents all information using natural language, which enables the direct use of any pretrained LLMs and VLMs without domain-specific knowledge or finetuning. By utilizing powerful pretrained VLMs for visual perception, LIMP can directly recognize actions from RGB videos in an open-ended way, without specifying target action labels for a domain. LIMP also leverages an LLM to use contextual clues from the text to fill in missing information from visual perception, providing a general method for multi-modal information fusion. One can also easily upgrade LIMP by plugging in \textit{any} state-of-the-art VLMs and LLMs. 

\textbf{What are the limitations of LIMP?} 
Hallucinations created by the VLM can cause significant errors in LIMP. For example, an agent may only open and close the fridge, but the VLM may mistakenly think that the agent also grabs something from the fridge. Such hallucinations in action recognition can not be corrected by the textural context. As a result, LIMP will incorrectly interpret the agent's behavior. Additionally, LIMP does not explicitly infer an agent's belief of another agent's belief. It instead prompts an LLM with past actions and utterances to implicitly infer that, which can become costly for longer events. LIMP also does not perform recursive reasoning for more than two levels.

\textbf{What are the limitations of our benchmark?}
The scenarios in our benchmark are currently limited to interactions between two agents in household settings, where there are three social goals: helping, hindering, and acting independently. Moreover, the current benchmark has synthetic human activities. These synthetic activities are realistic as verified in prior work \cite{puig2020watchandhelp} and enable large-scale testing. However sim-to-real evaluation could be valuable for future studies.

\section{Conclusion}
We present the first multi-modal Theory of Mind benchmark for multi-agent interactions in complex embodied settings. We have systematically evaluated humans and state-of-the-art LMMs on our benchmark. We have also proposed a novel multi-modal ToM model that outperforms all baselines while maintaining generality. In future work, we intend to incorporate more complex real-world scenarios beyond household environments and introduce multi-modal social interactions involving more than two agents. We also plan to create a test set with real-world videos for ToM evaluation in real-world scenarios.
\bibliography{aaai25}

\clearpage

\begin{strip}%
 \centering
 \LARGE \textbf{Appendix} \\[3em]
 \normalsize
\end{strip}

\section{Comparison of ToM Benchmarks}
\vspace{0.2cm}
\begin{table*}[htbp!]
  \begin{center}
    \begin{small}
    \begin{tabular}{p{2cm} p{1.5cm} p{2.5cm} p{0.5cm} p{1cm} p{2cm} p{2cm} p{2cm}}
    \toprule
      \textbf{Benchmark} & \textbf{Agent number} & \textbf{Tested concepts} & \textbf{Size} & \textbf{Modality} & \textbf{Communication} & \textbf{Generation} & \textbf{Evaluation}\\
    \hline
        \textbf{Triangle COPA \cite{gordon2016commonsense}}  & Single agent & Social Interaction & 100 & Text & No & Hand-designed & Multiple choice Q\&A  \\ \hline
        \textbf{ToMi \cite{le2019revisiting}} &  Multi agents & First \& Second Order belief & 400 & Text & No & Templates & Multiple choice Q\&A \\ \hline
        \textbf{Phase \cite{NetanyahuPHASE2021}} & Multi agents & Goals and Social relationships & 500 & Video & No & Procedural Generation & Multiple choice recognition \\ \hline
        \textbf{Agent \cite{shu2021agent} } & Single agent &  Goal Preferences, Action Efficiency, Unobserved Constraints, and Cost-Reward Trade-offs & 960 & Video & No & Procedural Generation & Surprise ratting \\ \hline
        \textbf{Epistemic reasoning \cite{cohen2021exploring}} & Multi agents & Knowledge and Belief & 2000 & Text & No & Templates & True or false judgements \\ \hline
        \textbf{BIB \cite{gandhi2021baby}} & Single \& Multi agents & Goal Preferences, rational actions, constraints & 5000 & Video & No & Procedural Generation & Surprise rating \\ \hline
        \textbf{Adv-CSFB \cite{kosinski2023theory}} & Single agent & False belief & 183 & Text & No & Hand-designed & Multiple choice filling in the blanks \\ \hline
        \textbf{Hi-ToM \cite{he2023hi}} & Multi agents & High-order beliefs & 600 & Text & Yes & Procedural Generation & Multiple choice Q\&A \\ \hline
        \textbf{FANToM \cite{kim2023fantom}} & Multi agents & Belief \& information tracking & 4807 & Text & Yes & Procedural Generation & Question answering \\ \hline
        \textbf{BigToM \cite{gandhi2024understanding}} & Single agent & Belief & 5000 & Text & No & Procedural generation & Question answering \\ \hline
        \textbf{MMTOM-QA \cite{jin2024mmtom}} & Single agent & Belief \& Goal & 600 & Text \& Video & No & Procedural generation & Multiple choice Q\&A \\ \hline
        \textbf{TomBench \cite{chen2024tombench}} & Multi agents & Emotion, desire, intention, knowledge, belief, non-literal communication & 5330 & Text & Yes & Procedural generation & Multiple choice Q\&A \\ \hline
        \textbf{OpenToM \cite{xu2024opentom}} & Multi agents & Second-order belief, attitude & 696 & Text & No & Procedural generation & Question answering \\ \hline
        \textbf{Negotiation ToM \cite{chan2024negotiationtom}} & Multi agents & Belief, desire, intention & 13800 & Text & Yes & Procedural generation & Question answering \\ \hline
        \textbf{Infant Cognition Benchmark \cite{LiAnII}} & Multi agents & False belief, social goal & 2000 & Video & No & Procedural generation & Surprise rating \\ \hline
        \textbf{Common-ToM \cite{soubki2024views}} & Multi agents & High order belief & 2104 & Text & Yes & Procedural generation & True of false judgements \\ \hline
        \textbf{EmoBench \cite{sabour2024emobenchevaluatingemotionalintelligence}} & Multi agents & Complex emotions, personal beliefs \& experiences, emotional cues, perspective taking & 200 & Text & Yes & Hand-designed & Multiple choice Q\&A \\ \hline
        \textbf{Our MuMA-ToM benchmark} & Multi agents & Belief, social goal and belief of other's goal & 900 & Text \& Video & Yes & Procedural generation & Multiple choice Q\&A \\
    \bottomrule \\
    \end{tabular}
    \end{small}
    \caption{Comparison between MuMA-ToM and prior ToM Benchmarks}
    \label{tab:comparison_table}
  \end{center}
\end{table*}

Table~\ref{tab:comparison_table} provides a comparison between our MuMA-ToM benchmark and prior ToM benchmarks, highlighting key features such as the size of the test set, input modalities, and evaluation metrics. Our benchmark stands out as the only benchmark with multi-modal inputs and multi-agent interactions. It simultaneously evaluates multi-agent social interactions with belief, goal, and belief of other agents' goals, as well as the ability to infer mental states from multi-modal inputs.

\section {MuMA-ToM Benchmark Details}
\vspace{0.3cm}

\subsection {More Quantitative Results}
\vspace{0.1cm}
 The results of all experiments conducted in our study are shown in Table~\ref{tab:results}. 
\subsubsection{Chain of Thought Prompting.} We evaluate state-of-the-art models' performance on our dataset with zero-shot chain of thought (CoT) prompting, as introduced by \cite{Kojima2022LargeLM}. We add the phrase ``Let's think step by step'' after the question prompt but before the list of options. 

For all models tested, using CoT prompting showed no significant improvement in performance. In fact, for many models, using CoT prompting caused a decrease in performance. While there are instances where CoT led to some improvement, such as in belief inference for InternVL 2 26B, the overall impact effect was negligible on more challenging social goal and belief of goal inference questions. These results further highlight the current limitations of state-of-the-art LMMs. Even with CoT guidance, they struggle to effectively understand social interactions.

\subsubsection{Finetuned Baseline}
We finetuned the VideoLlama 2 7B model on our training set for action captioning tasks following \cite{damonlpsg2023videollama}, using two A100 GPUs for 1 epoch, with a learning rate of 2e-5 and a batch size of 4. The performance of the model was lower after finetuning, suggesting that the model may have inherent limitations in ToM reasoning or action recognition. We experimented with finetuning for up to 3 epochs and found that extending finetuning beyond one epoch leads to over-fitting, and the model was unable to answer the questions with A, B, or C. 

\subsubsection{Advanced Prompting for ToM.}
Recent works have leveraged language models to tackle ToM problems through multi-step reasoning approaches \cite{wilf2023thinktwiceperspectivetakingimproves, sclar2023mindinglanguagemodelslack, hou2024timetom}. Among these text-only models, we chose to evaluate SimToM, as the code for the other models was either unavailable or required extensive modifications to integrate with our benchmark. Since SimToM only accepts textual input, we adapted it to our dataset by adding Gemini 1.5 Pro's visual extraction results after the textual input as input for SimToM and tested it with GPT-4o serving as the primary language model. SimToM, which analyzes the perspective of each agent to assist the language model, achieved the highest accuracy in belief-of-goal questions among all the baselines tested. This suggests that a multi-step approach can improve a language model's capacity for ToM reasoning. However, the overall accuracy is still below 50\%.

\subsubsection{LIMP w/ Llama 3.1 8B for Inverse Multi-agent Planning}
Solving ToM problems with language models usually requires some form of finetuning or few-shot prompting to equip the model with domain-specific knowledge. In contrast, LIMP leverages the forward planning capabilities of language models to address the inverse planning problem without any finetuning or additional domain knowledge. Beyond testing very large models like GPT-4o, we also explored the potential of smaller models, such as Llama 3.1 8B, as an inverse planner for LIMP. However, the results indicate that smaller models lack the ability to effectively function as inverse planners for multi-agent actions. A closer qualitative examination of Llama 8B's failure patterns shows that the model is unable to understand the concept of hindering, which leads to poor performance across all questions related to hindering.

\begin{table*}[t!]
  \begin{center}
    \begin{small}
    \begin{tabular}{l|c|c|c|c|c|c|c|c}
    \toprule
      \textbf{Method} &\textbf{Belief Inference} & \textbf{Social Goal Inference} & \textbf{Belief of Goal Inference} & \textbf{All}\\
    \hline
    Llava 1.6 34B & 93.6 & 37.2 & 27.5 & 52.8 \\
    Llava 1.6 34B CoT & 93.2 & 46.1 & 19.4 & 52.9 \\
    Llava 1.6 13B & 70.2 & 43.2 & 17.9 & 43.7 \\
    Llava 1.6 13B CoT & 64.9 & 41.6 & 25.3 & 43.9 \\
    Gemini 1.5 Flash & 53.9 & 33.0 & 41.4 & 42.7 \\
    Gemini 1.5 Flash CoT &  56.7 & 35.6 & 41.4 & 43.6 \\
    Gemini 1.5 Pro & 78.9 & 43.9 & 46.9 & 56.4 \\
    Gemini 1.5 Pro CoT & 79.8 & 42.6 & 41.1 & 54.5 \\
    GPT-4o & 67.9 & 39.6 & 44.4 & 50.6 \\
    GPT-4o CoT & 62.2 & 33.6 & 39.8 & 45.2\\ 
    InternVL 2 8B & 62.2 & 44.6 & 45.1 & 50.6 \\
    InternVL 2 8B CoT & 57.7 & 44.9 & 43.5 & 48.7 \\
    InternVL 2 26B & 59.3 & 44.9 & 35.5 & 46.6 \\
    InternVL 2 26B CoT & 64.1 & 44.9 & 36.1 & 48.4 \\ 
    VideoLlama 2 7B & 70.1 & 45.6 & 37.7 & 51.1 \\
    VideoLlama 2 7B CoT & 51.8 & 42.9 & 34.9 & 42.8 \\
    VideoLlama 2 7B (finetuned) & 42.7 & 35.7 & 34.3 & 37.3 \\
    SimToM  & 54.6 & 43.5 & 44.8 & 47.6 \\
    LIMP with Llama 3.1 8B & 35.8 & 23.4 & 37.7 & 33.0 \\
    BIP-ALM &41.2 & 34.1 & 30.6 & 33.9 \\
    LIMP with GPT-4o & 93.4 & 67.7 & 68.7 & 76.6 \\
    \bottomrule
    \end{tabular}
    \end{small}
    \caption{All experiment results: For models that accept video input, the full videos were provided. For models that do not, uniformly sampled frames (every 20 frames) were used instead. Since SimToM is a text-based model, we provided it with the action recognition outputs from Gemini 1.5 Pro.}
    \label{tab:results}
    \vspace{-10pt}
  \end{center}
\end{table*}

\subsection{Qualitative Results}
We provide two examples where Gemini 1.5 Pro, the best-performing LMM on the MuMA-ToM benchmark, fails while LIMP succeeds, highlighting the challenges state-of-the-art LMMs face on our benchmark. We also provide an example where hallucinations lead to LIMP also failing to solve the problem. \\

\noindent\textbf{Example of Gemini's due to failure to understand diverse social interactions}

\begin{figure}[h!]
    \centering
    \includegraphics[trim=0cm 8cm 17cm 0cm, clip, width=0.47\textwidth]{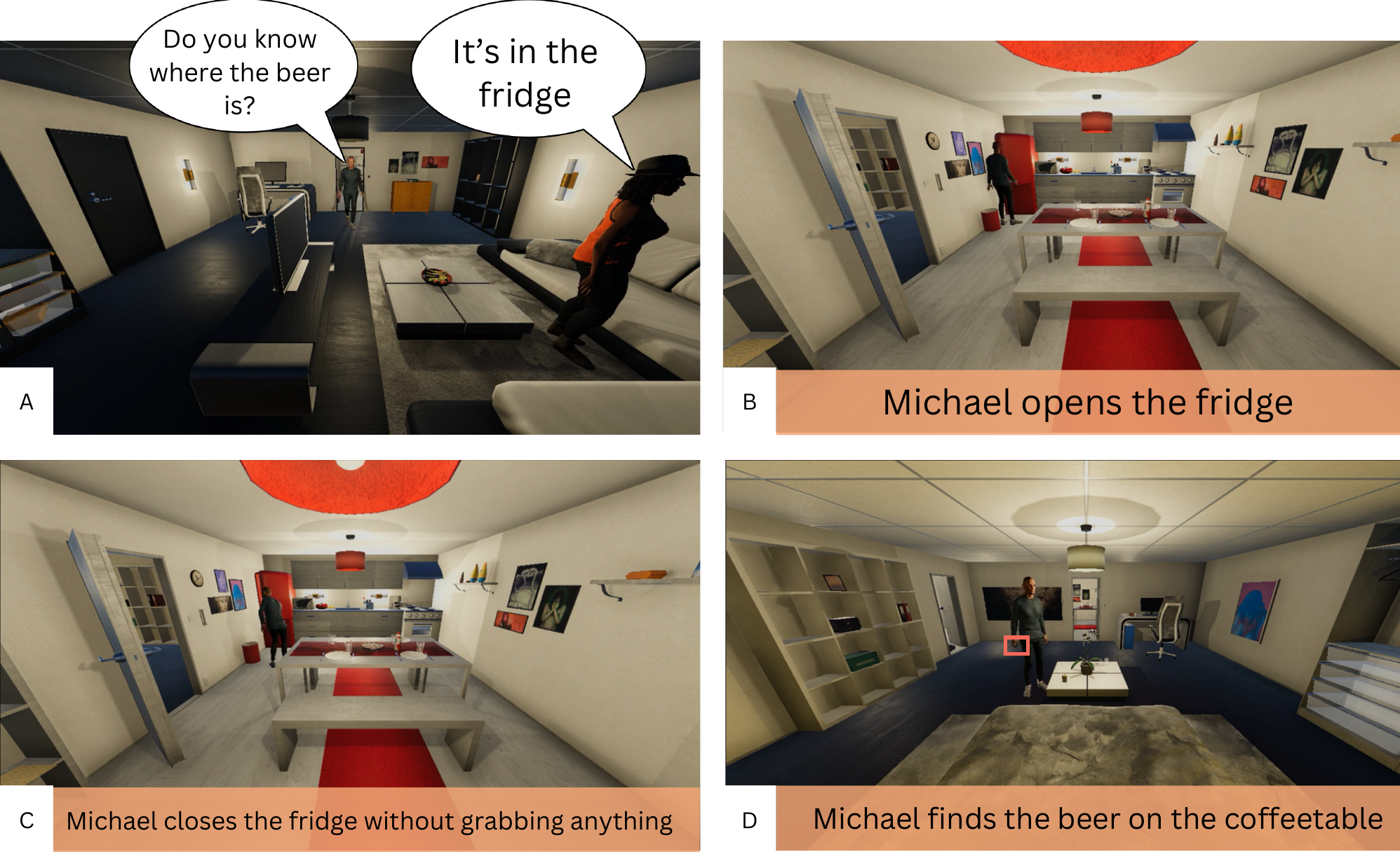}
   \vspace{-20pt}
    \label{fig:example_1}
\end{figure}

\begin{tcolorbox}

\small

\noindent
Question: Given the above interaction, assuming that Jessica knows what is inside the fridge, which of the following statements is LEAST likely? \\
A) When giving information, Jessica was indifferent towards Michael's goals. \\
B) When giving information, Jessica has been trying to prevent Michael from finding the beer. \\
C) When giving information, Jessica has been trying to help Michael locate the beer. \\
Correct Answer: C \\
Gemini Answer: B 
\end{tcolorbox}

The correct answer to this question is `helping' because it is assumed that Jessica knows there is no beer in the fridge, yet still directs Michael there, suggesting deliberate misleading, so helping is least likely. However, Gemini fails to correctly understand the concept of hindering. It assumes that as long as the agent provides information when asked, it is trying to help, regardless of whether the information is accurate. Since Gemini cannot interpret the act of providing misleading information as hindering, it fails to answer this question correctly. Gemini's explanation for its answer is shown below. \\

\noindent
\textbf{Gemini's answer and explanation:} \\
\texttt{The correct answer is \textbf{B)} When giving information, Jessica has been trying to prevent Michael from finding the beer.} \\
\texttt{Reason: We can see in the video, after Michael asks where the beer is, Jessica goes to the living room and waits there. When Michael goes to the kitchen, Jessica makes no attempt to stop him, this suggests that she is okay with him finding the beer.} \\
\texttt{Let's look at why the other answers are incorrect:} \\
\texttt{A) When giving information, Jessica was indifferent towards Michael's goals - This could be true as Jessica doesn't explicitly state that she wants him to find the beer nor does she show any disappointment when he finds it.} \\
\texttt{C) When giving information, Jessica has been trying to help Michael locate the beer - This is the most likely answer. Jessica provided the location of the beer to Michael after he asked, making it easier for him to find it.} \\ \\
\newpage
\noindent \textbf{Example of Gemini's failure due to visual recognition failure}

\begin{figure}[h!]
    \centering
    \includegraphics[trim=0cm 8cm 17cm 0cm, clip, width=0.47\textwidth]{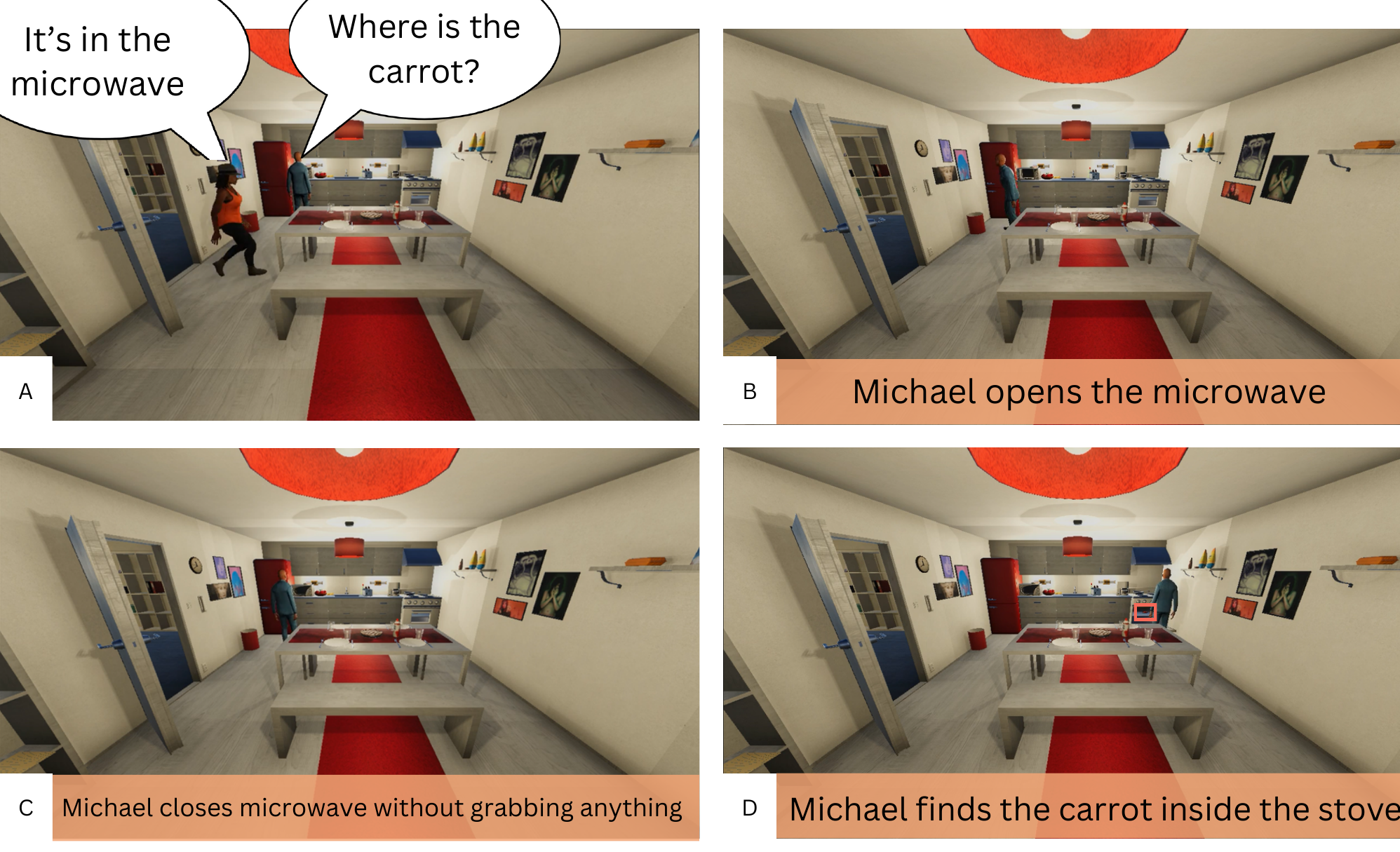}
   \vspace{-10pt}
    \label{fig:example_2}
\end{figure}

\begin{tcolorbox}

\noindent
Question: Given the above interaction, assuming that Mary knows what is inside the microwave, which of the following statements is MOST likely? \\
A) When giving information, Mary was indifferent towards Michael's goals \\
B) When giving information, Mary has been trying to help Michael locate the carrot \\
C) When giving information, Mary has been trying to prevent Michael from finding the carrot \\
Correct Answer: C \\
Gemini Answer: B
\vspace{10pt}
\end{tcolorbox} 

\noindent \textbf{Gemini's answer and explanation:}\\
\noindent \texttt{The answer is B) When giving information, Mary has been trying to help Michael locate the carrot.
Reasoning:
The dialogue clearly shows that Michael is seeking information about the carrot's location. Mary responded directly, indicating she understood his request and provided effective information. Even though Michael's further seek of carrot is not shown, this suggests she was trying to assist him.
There's no evidence to support options A or C. Mary's direct and relevant response implies she's not being indifferent or obstructive but rather cooperative.} \\

The correct answer to this question is `hindering' because it is assumed that Mary knows there is no carrot in the microwave, yet still directs Michael there, suggesting deliberate misleading. Gemini fails to recognize this because it cannot identify the carrot in the video. Without this recognition, it is unable to verify whether Michael actually found the carrot in the microwave and mistakenly interprets Mary's instructions as an attempt to help. \\ \\ 

\noindent
\textbf{Example of LIMP's failure due to hallucination}

\begin{figure}[h!]
    \centering
    \includegraphics[trim=0cm 7cm 16cm 0cm, clip, width=0.47\textwidth]{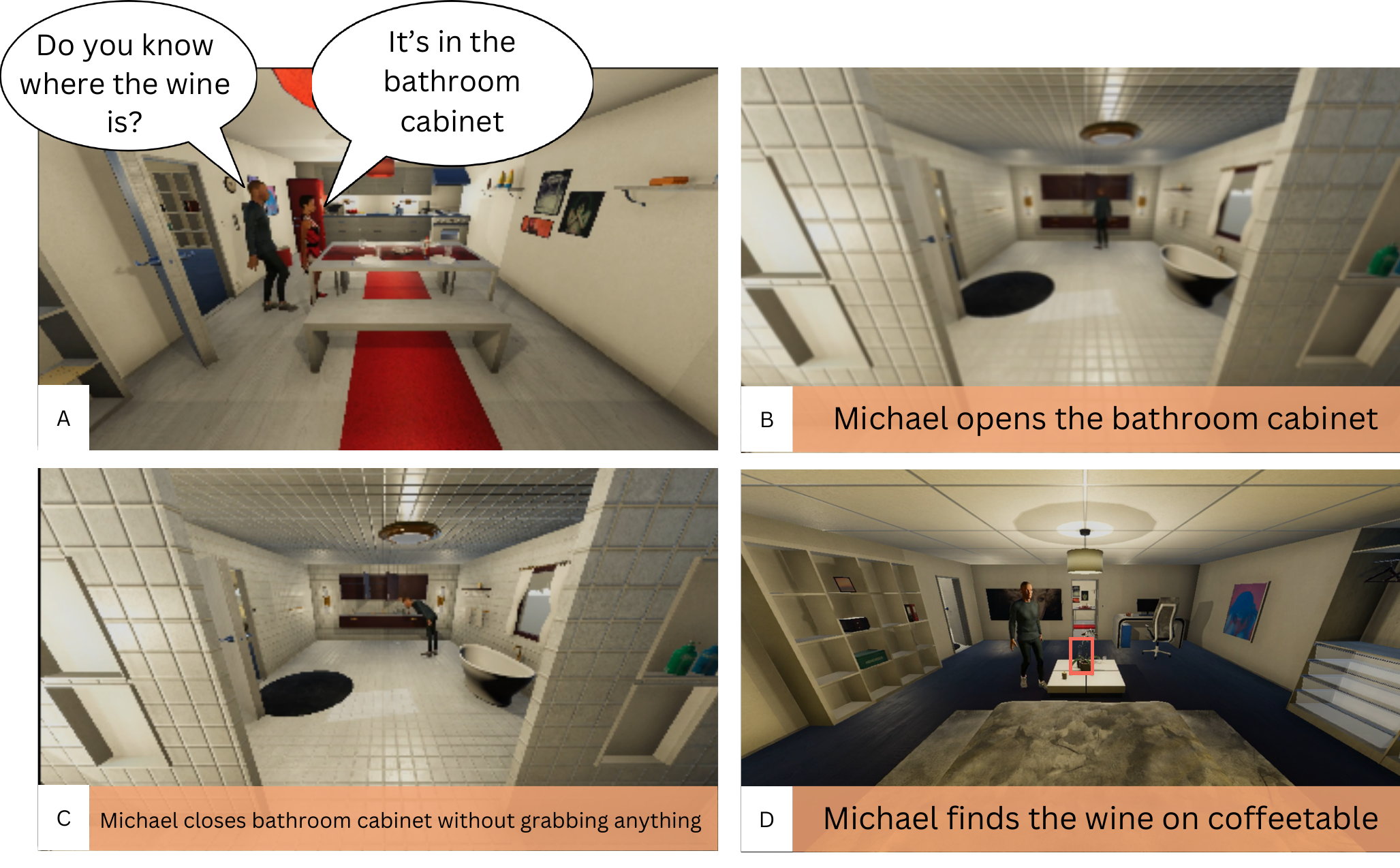}
   \vspace{-20pt}
    \label{fig:example_3}
\end{figure}

\begin{tcolorbox}
Question: Given the above interaction, assuming that Sarah knows what is inside the bathroom cabinet, which of the following statements is LEAST likely? \\
A) When giving information, Sarah was indifferent towards Michael's goals \\
B) When giving information, Sarah has been trying to prevent Michael from finding the wine \\
C) When giving information, Sarah has been trying to help Michael locate the wine \\
Correct Answer: C \\
LIMP's answer: B

\end{tcolorbox}

The question is similar to the previous two, with Sarah providing misleading information. Gemini 1.5 Pro here mistakenly extracts Michael grabbing the wine from the bathroom cabinet while Michael actually does not. As the initial state retrieval process in LIMP relies on the action, LIMP will mistakenly identify the location of wine as inside the bathroom cabinet and mistakenly interpret Sarah's misleading words as helping.

\subsection{Benchmark Statistics}
\begin{figure*}[t!]
    \centering
    \begin{minipage}{\textwidth}
        \includegraphics[trim=3cm 0cm 4cm 4cm, clip, width=\textwidth]{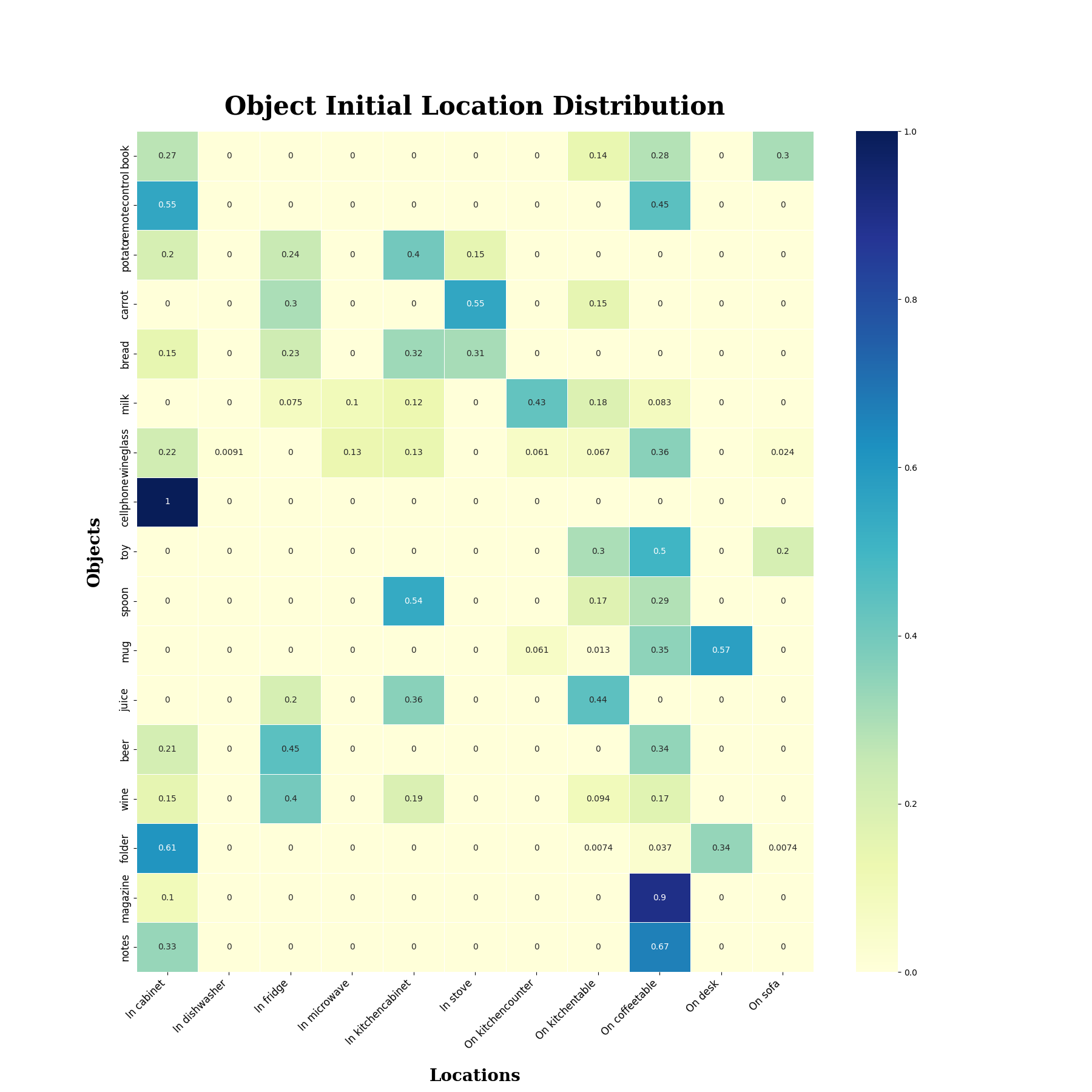} 
        \caption{Objects' initial locations in MuMA-ToM.}
        \label{object_locations}
    \end{minipage}
\end{figure*}

There are 225 interactive scenarios in our MuMA-ToM benchmark, 150 of which have language communication, and 75 of which do not have communication. The episodes in the benchmark are generated to be factually correct, concise, and human readable. Each interactive scenario happens in one of the four apartment with 10+ containers, 10+ surfaces, and 300+ objects in total. There are 17 relevant objects for the agents' goal in total, distributed among 11 initial locations. Figure \ref{object_locations} shows the distribution of objects' initial location.

\begin{figure*}[t!]
    \centering
    \begin{minipage}{\textwidth}
    \centering
        \includegraphics[trim=2cm 0cm 3cm 0cm, clip, width=0.7\textwidth]{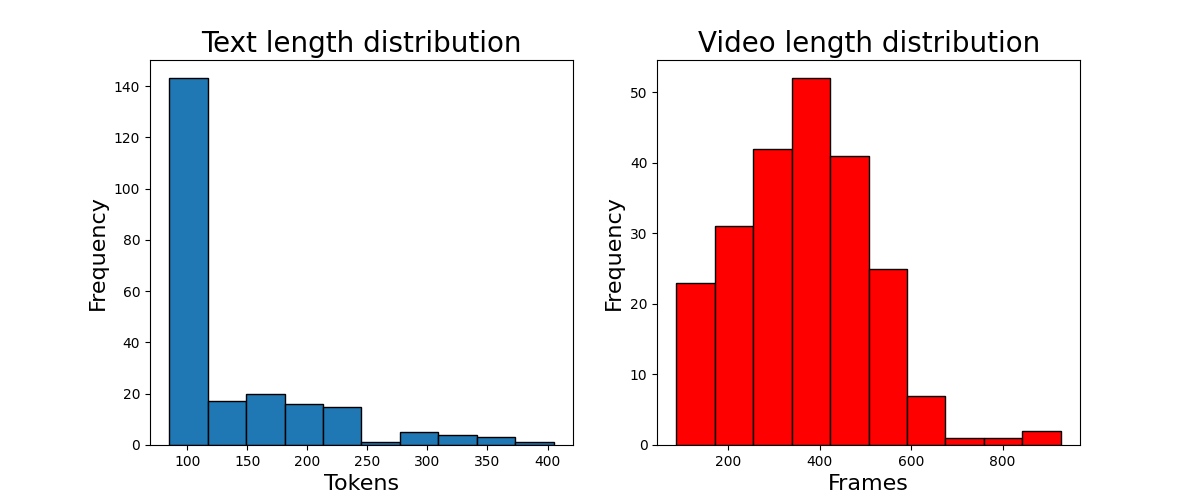} 
        \caption{MuMA-ToM context length distribution. The texts and videos are designed to be as concise as possible, allowing us to focus on testing models' ToM capability without long-context tracking. The videos are rendered at 10 frames per second.}
        \label{context_length}
        \vspace{-18pt}
    \end{minipage}
\end{figure*}

Figure \ref{context_length} shows the distribution of text and video length over all the scenarios. On average, the videos have 364.8 frames (approximately 36 seconds long), and the text inputs contain 136 tokens (many of which are just conversations). The relatively short context length reduces the need for the model to retrieve valid information from a large context, allowing us to focus on testing models' ToM capability without long-context tracking.

\subsection{Available Data}
We also provide depth images, instance segmentation, ground-truth actions, states, and camera data for our benchmark in addition to RGB videos and text. Even though our LIMP model does not rely on any of this information to make inferences, this information can be helpful for testing models' capability of solving ToM problems with some additional information: for example, ground-truth actions and object locations (from instance segmentation).

\subsection {Procedural Generation Details}
\begin{figure*}[htbp]
    \centering
    \begin{minipage}{\textwidth}
        \includegraphics[trim=0cm 0cm 0cm 0cm, clip, width=\textwidth]{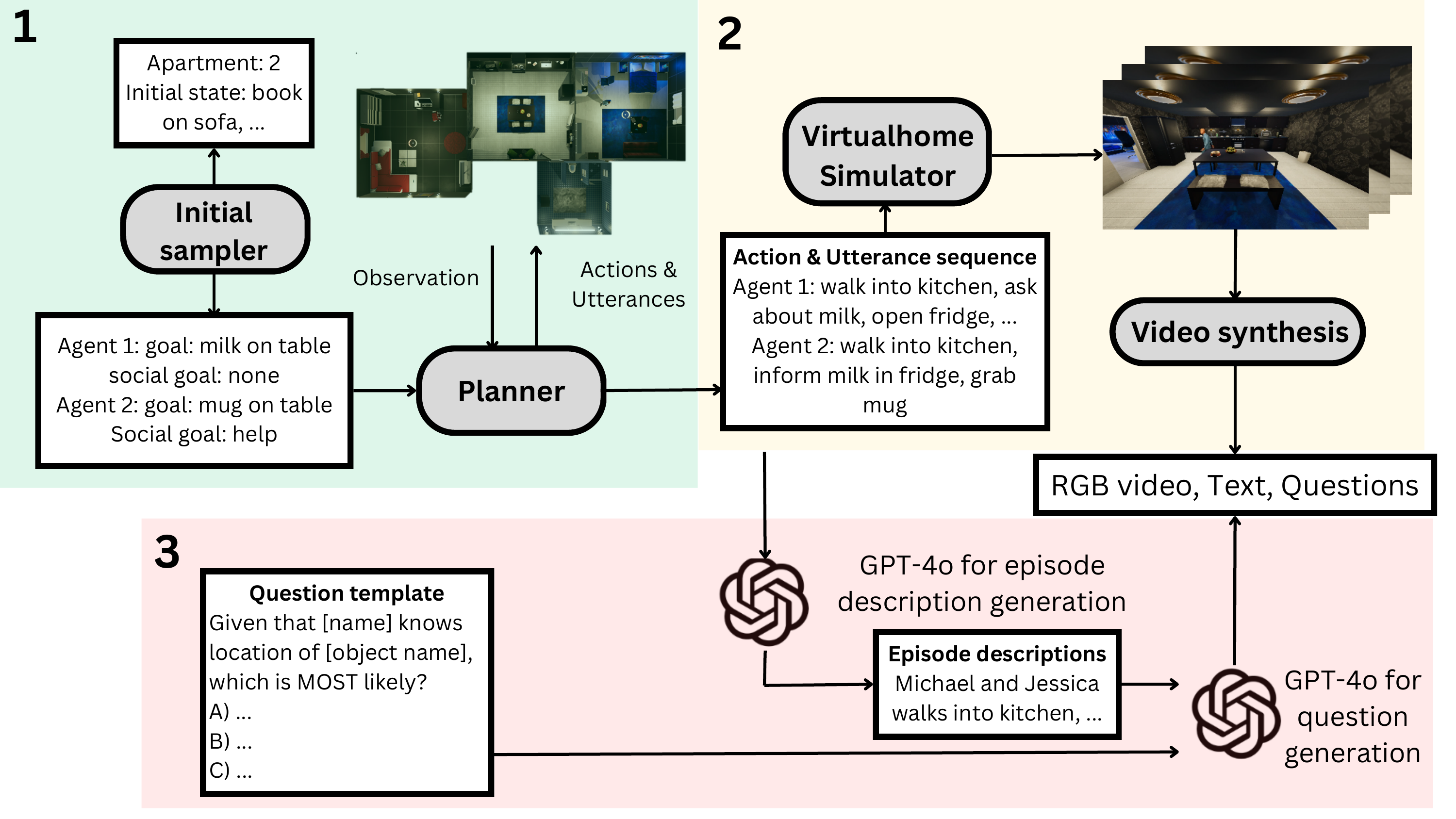} 
        \caption{Overview of the Procedural generation process. This method ensures that the episodes and ground truth answers are factually correct, while maintaining realistic conversations and scenarios.}
        \label{procedural_generation}
        \vspace{-10pt}
    \end{minipage}
\end{figure*}

Figure~\ref{procedural_generation} summarizes the procedural generation process. We follow a recent paper GOMA \cite{ying2024gomaproactiveembodiedcooperative} to generate actions \& utterance sequence, use the virtualhome \cite{puig2018virtualhomesimulatinghouseholdactivities} 3D simulator to generate humanoid actions within a realistic household environment and use GPT-4o to generate texts and questions.

Step 1 in Figure \ref{procedural_generation} shows the action \& utterance sequence generation process. We use four different apartments as the base environment for two agents' interactions, sampling objects to different containers \& surfaces within the apartment to generate a distinctive environment for each interactive scenario. Two agents' initial location (room location), physical goal (finding or rearranging an object), initial belief (ground-truth belief, false belief, or uniform belief), and social intentions (help, hinder, independent) are also sampled. For interactive scenarios without language, we sampled the environment and agents' goal in a way that ensures two agents' are aiming to put the same object to different locations and there is only one object of that type in the environment. In this way, agents will have to rearrange the object after the other agent has placed the object.
Afterward, a Monte Carlo Tree Search (MCTS) planner is used to compute the action sequence for each agent. The utterance is computed separately: for each step, if the two agents are in the same room and the first agent is uncertain about its goal object's location (entropy of its belief probability distribution exceeds a threshold), the first agent will send an inquiry. Upon receiving the inquiry, the second agent will answer based on its social intention (provide a contradictory answer with its belief when trying to hinder), and the first agent will update its belief accordingly. As agents' beliefs do not necessarily match the ground-truth state, the combination of intention with the ground-truth environment state is complicated: for instance, providing false information can be interpreted as trying to help but failing due to mistaken belief or deliberately trying to hinder. After the original utterance is generated, we use GPT-4o to add variety and improve the quality of language communication. The prompts we use are shown below. 

\begin{tcolorbox}
\textbf{Prompt for adding variety for inquiry}

Objective: Generate natural language from a language template. \\
User Input: Questions with a basic templated format in the form of "Where is X? Where is Y?" \\
Instructions: Convert this question into natural conversational language. Make it seem like everyday conversation. If the user asks about multiple objects, combine the objects into a single question.
\end{tcolorbox}

\begin{tcolorbox}
\textbf{Prompt for adding variety to response}

Objective: Generate natural language from language template. \\
User Input: The locations of an object with a basic templated format, with entries separated by ;. For instance, apple on table 121 livingroom; apple inside fridge 240 kitchen; apple null; banana on counter 101 kitchen; banana null means that there is an apple on the table in the livingroom, an apple inside the fridge in the kitchen, and the location of the third apple is unknown. There is a banana on the kitchen counter, and the location of the other banana is unknown. \\
Instructions:
Convert this statement into natural conversational language. If the multiple locations are provided for the same object, and some of them are null, ignore the null objects in the final description.
In the above example, ignore apple null and banana null since the locations of the other apples and banana are known.
\end{tcolorbox}

Step 2 in Figure \ref{procedural_generation} shows the visual generation procedure. After generating raw action \& utterance sequence, we use the Virtualhome simulator to render RGB, depth, and instance segmentation frames as well as supporting data like camera data or scene graphs. Then, raw frames are combined together into a video. For interactive scenarios with language, agent names', avatars and communication are overlaid as captions on the video frames. 

Step 3 in Figure \ref{procedural_generation} shows the text generation procedure for input text and questions. With two agents' actions and utterance sequence, we prompt GPT-4o to generate a description of the two agents' actions in a story-like way while maintaining chronological sequence. Portions of these descriptions are then used as textual input for the questions: for language scenarios, the conversation serves as the text input; for no-language scenarios, one agent's action is provided as the text input while the other agent's action is shown in the video. We then prompt GPT-4o to generate questions based on our pre-designed templates and the full descriptions. Essentially, GPT-4o fills in the blanks of the question templates using the information from the complete description. 

\begin{tcolorbox}
\textbf{Prompt for generating episode descriptions}
Objective: Create a description of a two-agent interaction scenario based on the provided language template. \\

\end{tcolorbox}
\begin{tcolorbox}
User Input: A list of actions by each agent, Verbal communication between the agents. \\
Structure: Actions: A list of actions taken by agent 0 and agent 1, Language: Verbal communication between the agents in a list format. \\
Instructions: \\
1. Synchronization guidelines: Synchronize actions and language, the first entry in the "language" list corresponds to the first action step, the second entry in the "language" list corresponds to the second action and so on. If a language entry is null, there is no communication at that timestep. Synchronize descriptions of actions and language strictly by timesteps. \\
2. Agent names: Choose from a predefined list of common names. \\
3. Description guidelines: Describe the actions and language of both agents together, step by step. Avoid adjectives and excessive descriptions. Do not skip any action or language steps. \\
4. After establishing the timeline, make the description shorter, more concise and flow a lot like a story. Do not skip any actions \\
5. Place more emphasis on the events immediately following the language conversation (if any) \\
6. When describing a action involved with grabbing objects, make sure to also include the original place of the object \\
\text{[Example input and generated scenario]}
\end{tcolorbox}
\begin{tcolorbox}
\textbf{Prompt for generating question templates}

Objective: Generate questions from the language template by filling in the blank. \\
User input: A description of the episode. \\
Expected output: Questions following this templated format, filling in the blanks, denoted by [] where necessary. \\
Template: \\
Question:
Given the above interaction, assuming that [second agent's name] knows what is [inside\\on a location that second agent mentioned in the communication], which of the following statement is MOST likely? \\
A) When giving information, [second agent's name] has been trying to help [first agent's name] locate [the object that second agent mentioned] \\
B) When giving information, [second agent's name] has been trying to prevent [first agent's name] from finding [the object that second agent mentioned] \\
C) When giving information, [second agent's name] was indifferent towards [first agent's name]'s goals \\
Answer: B) When giving information, [second agent's name] has been trying to prevent [first agent's name] from finding [the object that second agent mentioned]
\end{tcolorbox}

\section{LIMP Implementation Details}

\subsection{Visual Perception}
A previous multi-modal ToM model, BIP-ALM\cite{jin2024mmtom}, relies on instance segmentation to extract ground truth object information and spacial relationships from visual inputs. However, this level of detailed information is not available to other baselines or human participants, and is rarely available in real-world datasets. In RGB videos, detecting the exact object is challenging for both models and humans, as the objects might be small or obscured from view. To maintain generality, we use Gemini 1.5 Pro\cite{reid2024gemini} as our visual perception model. We upload the RGB videos to the Gemini web version in google AI studio without any extra information about possible locations and objects, and asked it to extract the action and utterance sequences of the agents. The prompt we use is shown below.

\begin{tcolorbox}
\textbf{Prompt for Gemini 1.5 Pro visual extraction}

Task: You will watch a video depicting two agents performing some actions. Your goal is to infer and describe the actions in chronological order. For [agent name], provide details about his/her actions, including what objects she handled, where she obtained them from, and where she placed them. Formulate all actions into a single line. Do not include any newline characters. Note that an agent moving their arm probably indicates opening a 
\end{tcolorbox}
\begin{tcolorbox}
container or picking up an item. If you cannot decipher the location that [agent name] grabs from, make your best guess
based on all the context in the video. If you cannot effectively identify the object, just leave it as \"grab some object\" without trying to guess the exact one.
\end{tcolorbox}

\subsection{Text Parsing \& Multi-modal Fusion}
For processing textual information, we directly use GPT-4o to parse the actions and utterances of each agent separately, in chronological order. Then, this parsed text information, along with the raw visual outputs from text input as well as raw visual outputs from Gemini, is provided to GPT-4o for information fusion.

A key step in our multi-modal fusion process we use is filling in missing information from the visual output based on the context. In the prompt given to Gemini, we instruct the model to leave blanks for exact object names, as accurately recognizing small or obscured objects is often impossible and could lead to unreasonable results. The raw visual output, along with text input that provides necessary context, is then used by GPT-4o to fill in these blanks with the correct object names mentioned in the context. This method reduces the model's reliance on recognizing small objects directly, and takes a more human-like approach to the problem.

Another important step in the multi-modal fusion process is initial state retrieval. The initial state of the environment is crucial for the planning process, as the agents' beliefs are based on the initial state instead of the changed state, unless they observe other agent moving things around directly. Since we do not use instance segmentation, it is challenging for the model to directly identify object locations or generate scene graphs from visual input. Instead, we use the agents' actions to infer the initial state of the environment. This reduces uncertainty for the model and allows it to focus on relevant objects to the interaction while ignoring unrelated ones.

The prompts we use for text parsing and multi-modal fusion are shown below.

\begin{tcolorbox}
\textbf{Prompt for text parsing}

You will read a piece of text describing actions of some number of people with distinctive names. You will also have a name, which is the name of the person whom you should pay attention to. Summarize the
person's actions and utterance separately
in a chronological order. Only include the actions and utterance directly taken by the person in the text, and exclude any 
\end{tcolorbox}

\begin{tcolorbox}
previous actions mentioned indirectly. If you cannot find either utterance or actions of the person in the text, leave the corresponding section blank. When reading words like "it", replace it with inferred object or location to make actions clearer. Do not include agent's communication as part of it. Organize your answer in this form:
Actions:
["action one", "action two", "action three", ...]
...
Utterance:
["utterance one", "utterance two", "utterance three", ...]
... 

Text: {text input}

Name: {name of the agent}
\end{tcolorbox}

\begin{tcolorbox}
\textbf{Prompt for error recovery}

You will read some text describing a person's action. The name of the person is given. Summarize and reorganize the person's actions.
Possible actions include walk towards somewhere, grab something from somewhere, open some container, close some container, put something somewhere. Only summarize these actions and their synonyms in this form and abandon mismatch actions. Omit person's name. When mentioning location name, try to infer room the location is inside and include it in the action
Check objects mentioned in the Additional Information section. Replace any object mentioned in action with the object appeared in that section
Formulate your final answer in the following form.
Actions:
["action1", "action2", ....] \\

Input text: {raw output of Gemini} \\
Additional information: {context} \\
Person's name: {agent name}

\end{tcolorbox}

\begin{tcolorbox}
\textbf{Prompt for initial state retrieval}

You will read one or two person's actions in a list like form. From the actions taken, extract the initial state of the environment before any people act. 
Check each grab action or synonyms. Describe it in the form "There is a [object grabbed] [on/inside location of grabbing].
Only include environment states statements. Do not include any other information or extra contents.

Actions: {all agents' actions}

\end{tcolorbox}

\subsection{Hypothesis Parsing}

We identify the three latent variables: belief, social goal and the belief of goal for understanding social interactions. The questions are designed in a way that for each option, there will be a set of these three latent variables corresponding to it. In the latent variable extraction stage, GPT-4 is prompted to extract the three sets. Initial state and actions of agents are also given as context as there are descriptions like "knows the location of the object" or "has put the object at desired location" requiring checking action \& initial state to figure out the exact location of the object. The prompt is shown below.

\begin{tcolorbox}
\textbf{Prompt for latent variable extraction}

You will read a question about agents' mind and ideas, and the initial state of the environment from which agents' are interacting in. Agents' knowledge \& belief are about this initial state, but not necessarily changed state after some actions. For each choice, extract one set of second person's belief (make sure to turn it into some statement about the environment state), second person's social goal toward first peron's actions (help, hinder or some similar words of indepedent), and second person's believed first person's physical goal (some arrangement of objects). Organize the answer in this way: A: Belief: contents; Social goal: contents; Believed Goal: contents. B: Belief: contents; Social goal: contents; Believed Goal: contents. C: Belief: contents; Social goal: contents; Believed Goal: contents. Do not include any other information or extra contents. Make sure your answer follow the format requirement, use ";" to separate variables within each choice and end response with ".". Separate contents of "A", "B" and "C" with "."

Initial state: {}

Question: {}

Actions: {}

\end{tcolorbox}

\subsection{Inverse Multi-agent Planning with GPT-4o}

Unlike open-source models, GPT-4o does not provide the log probability for any given completion, so the exact probability of the utterance or action cannot be calculated. However, GPT-4o does offer the log probabilities for the top 5 responses it generates. To address this, we implement a method that asks GPT-4o to assess the likelihood of a given utterance or action and restricts its most likely responses to two choices: A) Likely, or B) Unlikely. We then calculate the probability of the completion by using the log probability of the token 'A'. The prompt is shown below:

\begin{tcolorbox}
\textbf{Prompt for GPT-4o Inverse Planning}

Decide if agent's action is likely with the information provided, and respond with only either A or B:

{agent}'s social goal: {social goal}

{agent}'s belief: {belief}

{agent}'s belief of {other agent}'s goal: {belief of goal}

{other agent's utterance}: {utterance}

Initial state: {initial state}
\end{tcolorbox}

\begin{tcolorbox}

Previous Actions: {actions taken previously}

Respond with only either A or B:

Agent's Action or Utterance:

A) Likely

B) Unlikely

\end{tcolorbox}

\section {Baseline Implementation Details}

\subsection{LMMs}
For the large multi-modal models we tested on our benchmark, we provided the same videos for humans, LIMP, and state-of-the-art LMMs capable of handling video input. For Intern-VL, we use a provided script to sample 8-10 frames from our video, as it cannot directly handle video input. For other LMMs that only accept images as input, we uniformly sampled one frame every 20 frames (approximately every 2 seconds), resulting in 15-30 frames per video. For videos with language communication, additional context was provided regarding the timing of the communication to ensure the utterances align with the frame sequence. Specifically, for Llava 1.6 and GPT-4o, we provided frames as input. For Gemini 1.5 and VideoLlama 2, we provided the full videos. For InternVL 2, we used the officially provided script to sample 8-10 frames, adapting the video input to the model's frame-based capabilities.

\subsection{BIP-ALM}
To adapt BIP-ALM to work on our benchmark, we needed to provide additional information. Since BIP-ALM relies on instance segmentation and depth images for visual perception, we followed its method of projecting pixels to 3D and detecting object relationships based on bounding boxes \& instance segmentations. We also enhanced the original object detection mechanisms (which only detected closeness, inside and open relationships) by adding the detection of 'on' and 'holding' relationships. Additionally, because BIP-ALM uses symbolic representations, its text parsing module requires a strict data format. Therefore, we provided descriptions of the full interactive scenarios to ensure BIP-ALM functions correctly. Despite receiving more information than LIMP, BIP-ALM still performed no better than random guessing on MuMA-ToM, suggesting that single-agent inverse planning models cannot be directly applied to multi-agent interactions.

\section {Full Version of the Example Questions}
\subsection{Belief Inference}
Video: \url{https://youtu.be/kXCUL7KB2Dc} \\
\begin{tcolorbox}
\textbf{Text: }
John: Do you know where the beer is? \\
Mary: I discovered a beer on the coffee table in the living room. \\

\textbf{Question:} Given the above interaction, if Mary has been trying to hinder John from achieving his goal, which of the following statements is LEAST likely? \\
A) When giving information, Mary believed that there was beer on the coffee table in the living room \\
B) When giving information, Mary believed that there was beer inside the kitchen cabinet \\
C) When giving information, Mary believed that there was beer inside the fridge \\

\textbf{Correct Answer: } A

\end{tcolorbox}

\subsection{Social Goal Inference}

Video: \url{https://youtu.be/e4360D7yYV8} \\
\begin{tcolorbox}
\textbf{Text: }
Kevin: Any idea where the magazine might be? \\
Jessica: I discovered a magazine inside the cabinet in the bedroom. \\

\textbf{Question: }Given the above interaction, assuming that Jessica knows what is inside the cabinet in the bedroom, which of the following statements is MOST likely? \\
A) When giving information, Jessica has been trying to help Kevin locate the magazine \\
B) When giving information, Jessica has been trying to prevent Kevin from finding the magazine \\
C) When giving information, Jessica was indifferent towards Kevin's goals \\

\textbf{Correct Answer: } B

\end{tcolorbox}

\subsection{Belief of Goal Inference}

Video: \url{https://youtu.be/bpw1jpbViFE} \\
\begin{tcolorbox}
\textbf{Text: }
David walked to a book, grabbed it along with another book. He then walked to the living room, headed to the bedroom, and finally reached the desk there, placing both books on the desk. \\
\textbf{Question: }"Given the above interaction, based on the actions of the agents, which of the following statements is MOST likely? \\
\end{tcolorbox}
\begin{tcolorbox}
A) Sarah believed that David placed the book at his desired location: she moved the book to the coffee table to help David. \\
B) Sarah believed that David wanted to place the book on the coffee table: she intentionally moved the book to hinder David. \\ 
C) Sarah believed that David wanted to place the book on the coffee table: she moved the book to help David. \\

\textbf{Correct Answer: } C

\end{tcolorbox}

\end{document}